\documentclass[11pt,a4paper]{article}
\usepackage[hyperref]{acl2021}
\usepackage{times}
\usepackage{latexsym}

\usepackage{microtype}

\aclfinalcopy % Uncomment this line for the final submission
\title{Challenges and Applications of Automated Extraction of Socio-political Events from Text (CASE 2021): Workshop and Shared Task Report}

\author{Ali Hürriyetoğlu \\
  Koç University \\
  Sarıyer, İstanbul, Turkey \\
  \texttt{\small ahurriyetoglu@ku.edu.tr} \\\And
  Hristo Tanev \\
  European Commission \\
  Ispra, Varese, Italy \\
  \texttt{\small hristo.tanev@ec.europa.eu} \\\And
  Vanni Zavarella \\
  European Commission \\
  Ispra, Varese, Italy \\
  \texttt{\small vanni.zavarella@ec.europa.eu} \\\AND 
  Jakub Piskorski \\
  Polish Academy of Sciences \\
  Warsaw, Poland \\
  \texttt{\small jpiskorski@gmail.com} \\\And
  Reyyan Yeniterzi \\
  Sabancı University \\
  Tuzla, İstanbul, Turkey \\
  \texttt{\small reyyan@sabanciuniv.edu} \\\And
  Erdem Yörük \\
  Koç University \\
  Sarıyer, İstanbul, Turkey \\
  \texttt{\small eryoruk@ku.edu.tr} \\
  \vspace*{250px}
}
  
\date{}

\begin{document}

\maketitle
\vspace*{86px}
\begin{abstract}
This workshop is the fourth issue of a series of workshops on automatic extraction of socio-political events from news, organized by the Emerging Market Welfare Project, with the support of the Joint Research Centre of the European Commission and with contributions from many other prominent scholars in this field. The purpose of this series of workshops is to foster research and development of reliable, valid, robust, and practical solutions for automatically detecting descriptions of socio-political events, such as protests, riots, wars and armed conflicts, in text streams. 
This year workshop contributors make use of the state-of-the-art NLP technologies, such as Deep Learning, Word Embeddings and Transformers and cover a wide range of topics from text classification to news bias detection. Around 40 teams have registered and 15 teams contributed to three tasks that are i) multilingual protest news detection, ii) fine-grained classification of socio-political events, and iii) discovering Black Lives Matter protest events. The workshop also highlights two keynote  and four invited talks about various aspects of creating event data sets and multi- and cross-lingual machine learning in few- and zero-shot settings.
\end{abstract}

\vspace*{84.5px}

\section{Introduction}

Today, the unprecedented quantity of easily accessible data on social, political, and economic processes offers ground-breaking potential in guiding data-driven analysis in social and human sciences and in influencing policy-making processes. The need for precise and high-quality information about a wide variety of events ranging from political violence, environmental catastrophes, and conflict to international economic and health crises has rapidly escalated~\citep{Dellaporta+2015,Coleman+2014}. Governments, multilateral organizations, and local and global NGOs present an increasing demand for this data to prevent or resolve conflicts, provide relief for those that are afflicted, or improve the lives of and protect citizens in a variety of ways. For instance, Black Lives Matter protests~\footnote{\url{http://protestmap.raceandpolicing.com}, accessed on June 2, 2021.}, conflict in Syria~\footnote{\url{https://www.cartercenter.org/peace/conflict_resolution/syria-conflict-resolution.html}, accessed on June 2, 2021.} and COVID-19 related events~\footnote{\url{https://en.wikipedia.org/wiki/Protests_over_responses_to_the_COVID-19_pandemic}, accessed on June 2, 2021.} are only a few examples where we must understand, analyze, and improve the real-life situations using such data.

A recent report from ReliefWeb~\footnote{\url{https://reliefweb.int/report/world/trends-armed-conflict-1946-2017}, accessed on June 3, 2021.} clearly demonstrates that the number of wars and other armed conflicts is on an increasing trend. In particular, the so-called internationalized conflicts are on a rise in the last two decades. In this situation, it is important to provide solutions for situation awareness, using various branches of artificial intelligence (AI), natural language processing (NLP), machine learning (ML), and advanced statistical methods.

In this clue, event detection and extraction plays an important role, because of its capacity to detect conflict developments in news and social media and to extract important information about them. Such information involves the quantity and the profiles of the victims, the participating entities, the conflict dynamics, its spatio-temporal characteristics, the weaponry used, as well as infrastructural, technical and human impact. This information extracted through various NLP methods can throw light on the intensity and the trend development of each conflict, as it is reflected in the media. Event detection has been used by political analysts to write their daily situation reports for decision makers, to create long-term analyses, as well as for conflict forecasting and prediction.

Automation offers scholars not only the opportunity to improve existing practices, but also to vastly expand the scope of data that can be collected and studied, thus potentially opening up new research frontiers within the field of socio-political events, such as political violence and social movements. Event information collection has long been a challenge for the NLP community as it requires sophisticated methods in defining event ontologies, creating language resources, and developing algorithmic approaches~\citep{Pustejovsky+2003,Tanev+08,Boros2018,Chen+2021}. We believe that this workshop and the shared task contribute strongly towards putting emphasis on this important technology, providing a gathering point for scientists and developers in NLP, AI, conflict studies and related areas.

Social and political scientists have been creating event databases such as ACLED~\citep{Raleigh+10}, EMBERS~\citep{Saraf+2016}, GDELT~\citep{Leetaru+13}, ICEWS~\citep{OBrien2010}, MMAD~\citep{Weidman+19}, PHOENIX, POLDEM~\citep{Kriesi+2019}, SPEED~\citep{Nardulli+15}, TERRIER~\citep{Liang+18}, and UCDP~\citep{Sundberg+2012} for decades. These projects and the new ones increasingly rely on machine learning (ML) and NLP methods to deal better with the vast amount and variety of data in this domain~\citep{Hurriyetoglu+2021b}. Nonetheless, automated approaches suffer from major issues like bias, low generalizability, class imbalance, training data limitations, ethical issues, and lack of recall quantification which affect the quality of the results and their use drastically~\citep{Leins+2020,Bhatia+2020,Chang+2019,Yoruk+2021}. Moreover, the results of the automated systems for socio-political event information collection may not be comparable to each other or not of sufficient quality~\citep{Wang+16,Schrodt2020}. 

Socio-political events are varied and nuanced. Both the political context and the local language used may affect whether and how they are reported. Therefore, all steps of information collection (event definition, language resources, and manual or algorithmic steps) may need to be constantly updated. This leads us to a  series of challenging questions such as: Do events related to minority groups are represented well? Are new types of events covered? Are the event definitions and their operationalization comparable across systems? We organize the workshop on Challenges and Applications of Automated Extraction of Socio-political Events from Text (CASE 2021)~\footnote{\url{https://emw.ku.edu.tr/case-2021/}, accessed on June 9, 2021.} and the shared task Socio-political and Crisis Events Detection~\footnote{\url{https://github.com/emerging-welfare/case-2021-shared-task}, June 12, 2021.} to seek answers to these and related questions, to inspire innovative technological and scientific solutions for tackling the aforementioned issues, and to quantify the quality of the automated event extraction systems. Moreover, the workshop aims to trigger a deeper understanding of the performance of the computational tools used and the usability of the resulting socio-political event datasets. The workshop is co-located with the Joint Conference of the 59th Annual Meeting of the Association for Computational Linguistics and the 11th International Joint Conference on Natural Language Processing (ACL-IJCNLP 2021). 

We invited contributions from researchers in computer science, NLP, ML, AI, socio-political sciences, conflict analysis and forecasting, peace studies, as well as computational social science scholars involved in the collection and utilization of socio-political event data. Social and political scientists are interested in reporting and discussing their approaches and observe what the state-of-the-art text processing systems can achieve for their domain. Computational scholars have the opportunity to illustrate the capacity of their approaches in this domain and benefit from being challenged by real-world use cases. Academic workshops specific to tackling event information in general or for analyzing text in specific domains such as health, law, finance, and biomedical sciences have significantly accelerated progress in these topics and fields, respectively. However, there is not a comparable effort for handling socio-political events. We hope to fill this gap and contribute to social and political sciences in a similar spirit. We invite work on all aspects of automated coding of socio-political events from mono- or multi-lingual text sources. This includes (but is not limited to) the broad topics below.

\begin{description}
 \item[Data:] collecting and annotating data, identifying the qualities, bias and fairness of the sources, handling ethics, misinformation, privacy, and fairness concerns pertaining to event datasets, respecting copyright of the sources at  the creation, dissemination, and release phases of an event dataset;
 \item[Task:] defining, populating, and facilitating event schemas and ontologies,  extracting events in and beyond a sentence, detecting event coreference and event-event relations such as subevents, main events, and causal relations, investigating lexical, syntactic, and pragmatic aspects of event information manifestation, determining socio-poltical events pertaining to societal issues such as COVID-19 and BLM, detecting novel events;
  \item[Approaches:] developing rule-based, machine learning, hybrid, and human-in-the-loop approaches for creating event datasets; and
 \item[Evaluation: ] evaluating event  datasets in light of reliability and validity metrics, estimating what is missing in event datasets using internal and external information, utilizing event datasets, releasing of new event datasets.
\end{description}

%online social movements, bias and fairness of the sources and event datasets, estimating what is missing in event datasets using internal and external information, novel event detection, release of new event datasets, ethics, misinformation, privacy, and fairness concerns pertaining to event datasets, copyright issues on event dataset creation, dissemination, and sharing, qualities of the event information on various online and offline platforms.

We provide summaries of the accepted papers, the shared task, keynotes, and invited talks in the sections \ref{acceptedpapers}, \ref{sharedtask}, \ref{keynotes}, and \ref{inviteds} respectively. Section~\ref{conclusion} concludes this report with main lessons derived from these efforts and interactions. 

\section{Accepted Papers}
\label{acceptedpapers}

%deVroe+2021 (submission no: 6), Raza+2021 (submission no: 9), Caselli+2021 (submission no: 10), Ramrakhiyani+2021 (submission no: 15), Radford+2021 (submission no: 19), Scharf+2021 (submission no: 20), Kar+2021 (submission no: 21)

The workshop attracted 21 submissions. The competition was high and 7 of them were accepted based on reviewer evaluations, which vary between 4 and 6 for each paper.

Here are brief descriptions of all accepted papers, except from the ones participating in the shared task, which are described in other papers in this proceedings:

\citet{deVroe+2021} present an open domain, lexicon-based event extraction system that captures various types of event modality. The definition of ``event'' in this work is quite broad, i.e. every predicate construction is taken into consideration. The authors use syntactic parsing to detect the event modality, which is a very important phenomena when making distinction between current, past and just probable events. The system explores conditionality, counterfactuality, negation, and propositional attitude. The achieved accuracy in the modality labelling task is 0.81 F1 that is measured on a small corpus of 100 manually annotated predicates.

\citet{Raza+2021} explores the topic of detecting fake news, which is potentially related to the trustability of the sources, from which events are extracted. The main approach in this study is based on a modified version of a pre-trained Bidirectional Encoder
Representations from Transformers (BERT) with the capability to receive as input news-related and side information. In particular, each news item is represented by its title (main
information) and side information, such as temporal, news-related information, author and source, as well as social contexts (related tweets) which give information about users’ reactions on the news. The proposed model is quite promising, considering it outperforms all other state-of-the-art methods. It achieves 96\% accuracy in deciding between fake and real news on a test set with fake and real news nearly equally represented.

\citet{Caselli+2021} explore how efficiently a retrained BERT model detects protest events. Authors present the PROTEST-ER system, which uses a retrained BERT model for protest event extraction. They use annotated event data from the protest event detection task following the 2019 CLEF ProtestNews Lab~\citep{Hurriyetoglu+19a,Hurriyetoglu+19b}. A worth-to-mention finding of this work is that PROTEST-ER outperforms a corresponding generic BERT with 8.1 points.

\citet{Ramrakhiyani+2021} describe a deep learning approach for detecting incidents from industrial reports. Incidents in industries have huge social and political impacts. However, automated analysis of repositories of incident reports has remained a challenge. Due to absence of event annotated datasets for industrial incidents authors employ a transfer learning based approach. A detailed analysis is provided on how amount of data utilized affect pre-training and why pre-training improves the performance. Data is gathered from aviation and construction incident reports. Different deep learning methods are evaluated for the task, including BiLSTM and transfer learning. Transfer learning consistently outperforms the baseline and achieves F1 measure of 0.81.

\citet{Radford+2021} presents a study on geocoding and a new data set. Geocoding is an important sub-task of event detection, in which the goal is to find the geographic coordinates associated with event descriptions. The paper presents an ``end-to-end probabilistic model'' for geocoding from text data. A novel data set has been created for evaluating the performance of geocoding systems. The output of the new model is compared with a state-of-the-art model, called Mordecai. The comparison clearly shows an improvement provided by the proposed model.

\citet{Scharf+2021} report on a study on the political bias in Hong Kong published news reporting about protest events. The paper reports on lexical differences between home and Western news sources about protests happening in Hong Kong in the period 1998-2020. Experiments on topic modeling, sentiment analysis, lexical distribution and comparative lexical analysis between Western- and Hong Kong-based sources reveal a bias in the reporting from the Hong Kong press. The evaluation reveals that during the Anti-Extradition Law Amendment Bill Movement reports from Hong Kong made fewer references to police violence compared to the Western media. The study also reveals that the lexical contexts of salient keywords changed in Hong Kong sources when the Movement emerged.

\citet{Kar+2021} describe an algorithm for event argument detection and aggregation. The paper reports on document level aggregation of the following argument types:
Time, Place, Casualties, After-Effect, Reason, and Participant. The ArgFuse algorithm is based on a BERT based active learning classifier, which identifies whether a pair of event arguments is redundant, and a Biased Text Rank argument ordering function. Authors report F1 measure of 0.61, which beats all the other 5 baseline algorithms with which the ArgFuse performance is compared.

\section{Shared Task: Socio-political and Crisis Events Detection}
\label{sharedtask}

The work on event database creation comprises of three steps that are collecting events, classifying them, and measuring utility of the system output, which is an event database, against ground-truth. Each of these steps contains pitfalls and subject to limitations. For instance, the data source utilized maybe biased or a ground-truth may not be available. Although aforementioned issues in socio-political and crisis event studies have been studied by numerous scholars for decades to date, there are still no answers or solutions to them~\cite{Wang+16,Lorenzini+16,Schrodt2020,Raleigh2020,Eck2021,Boschee2021}. Therefore, we aim at contributing to the understanding and resolution of event database creation via quantifying performance of the state-of-the-art text processing systems in the shared task Socio-political and Crisis Events Detection.~\footnote{~\url{https://github.com/emerging-welfare/case-2021-shared-task}, accessed on June 9, 2021.} 

The shared task consists of three tasks that are on collection (Task 1), classification (Task 2)~\citep{Haneczok+2021}, and evaluation (Task 3) of event databases. Shared task and submission details are reported in the overview papers of the tasks~\citep{Hurriyetoglu+2021c,Haneczok+2021,Giorgi+2021} and the system description papers in this proceedings respectively. We provide a summary of the tasks and the findings in the following subsections.

\subsection{Task 1: Multilingual protest news detection} 

The task is designed to be both multilingual (having both training and test data in English, Portuguese, and Spanish) and cross-lingual (having data in Hindi only for test). There are four subtasks that are document classification (subtask 1), sentence classification (subtask 2), event sentence classification (subtask 3), and event extraction (subtask 4). Event information is at the center of all of the subtasks, i.e. documents and sentences are classified as containing event information in subtasks 1 and 2, sentences that are about the same event are identified in subtask 3, and event trigger and its arguments are extracted in subtask 4. 

13 teams have submitted 238 submissions for the evaluation scenarios specified with subtask and language combinations. The best submissions utilized deep learning approaches that combine the training data in various languages, utilize large models, further re-train the models, and create ensemble models~\citep{Awasthy+2021,Hettiarachchi+2021,Re+2021,Hu+2021,Tan+2021}. Although training data was limited in Portuguese and Spanish and not available in Hindi, the best performing participants managed to deliver predictions that are between 77.27 and 93.03 F1-macro in subtasks 1, 2, and 3 for all languages. The performance of the best system for subtask 4 for all languages was between 66.20 and 78.11 for all languages and 4-5 F1-macro points ahead of all other teams in all languages.

\subsection{Task 2: Fine-grained Event Classification in News-like Text Snippets} 

Task 2 aims at evaluating conventional and generalized zero-shot learning event classification approaches to classify short text snippets reporting socio-political and crisis events. The task is divided into three subtasks: (a) classification of text snippets reporting socio-political events, using 25 events classes from the Armed Conflict Location and Event Data Project (ACLED) event taxonomy\cite{Raleigh+10}, for which vast amount of training data exists, although exhibiting slightly different structure and style vis-a-vis test data, (b) enhancement to a generalized zero-shot learning problem, where 3 additional event types were introduced in advance, but without any training data (`unseen' classes), and (c) further extension, which introduced 2 additional event types, announced shortly prior to the evaluation phase. Task 2 focuses on classification of events in English texts and the event definitions of events in this task are not fully compatible with those in Task 1.

8 teams registered, out of which 4 returned system responses, for Task 2. Best performing systems  for  the  subtask 1, 2 and 3 achieved 83.9\%,  79.7\% and 77.1\% weighted $F_{1}$ scores respectively. Most of the solutions submitted are built on top of fine-tuned Transformer-based models like {\sc Bert} and {\sc RoBERTa}. Given the specific set up of this task, i.e., the training data being some-what different from the test data and inclusion of some unseen classes the top results obtained can be considered good, however, there is place for improvement.

\subsection{Task 3: Discovering Black Lives Matter events in United States} 

Task 3 is only an evaluation task where the participants of Task 1 have the possibility to evaluate their systems on reproducing a manually curated Black Lives Matter (BLM) related protest event list. Participants use document collections, provided by the organizers and different from the documents from where Gold Standard has been extracted,  to extract place and date of the BLM events in these collections. The event definition applied for determining these events is the same as the one facilitated for task 1. Participants may utilize any other data source to improve performance of their submissions. The goal of the task is to achieve as high correlation as possible with the events from the Gold Standard, as computed by aggregating events on a regular cell geographical grid..

5 teams that performed the best in Task 1 were invited to participate in this task. In general all participating systems showed low levels of correlation with the Gold standard data, including the baseline system. The low recall at this year's shared task is most probably due to the low coverage of the test corpus, which participating systems have used, and its poor overlapping with manually collected Gold Standard Data. Two systems showed a relatively good performance: NoConflict \citet{Hu+2021} and EventMiner \citet{Hettiarachchi+2021}. The main lesson from this task is that Gold Standard data and test data should be checked for consistency and correlation. Moreover, this evaluation task highlighted some of the current limits on the usability of automatically extracted event datasets for modelling socio-political processes, such as fine-grained geocoding of events.

\section{Keynotes}
\label{keynotes}

Kristine Eck and Elizabeth Boschee will deliver the keynote talks. \citet{Eck2021} addresses the responsibility of the scholars that create event datasets to define and apply what is right, suggests data sources alternative to news data that may report event information inconsistently, and emphasizes the need for interdisciplinary collaboration for creating data sets that advance conflict studies. \citet{Boschee+13,Boschee2021} share the concerns addressed by Prof. Eck and presents a detailed study that compares various approaches for utilizing multilingual data in a cross-lingual zero-shot setting to improve quality of the event datasets.

\section{Invited Talks}
\label{inviteds}

The workshop contains an invited talks session as well. The authors of the papers published in Findings of ACL and related to workshop theme are invited to present their work in this session. The papers are 
\begin{description}
\item[\citet{Zhou+2021}] propose an event-driven trading strategy that predicts stock movements by detecting corporate events from news articles;
\item[\citet{Halterman+2021b}] introduce the IndiaPoliceEvents Corpus---all 21,391 sentences from  1,257 \textit{Times of India} articles about events in the state of Gujarat during March 2002;
\item[\citet{Halterman+2021a}] show utility of ``upsampling'' coarse document labels to fine-grained labels or spans for protest size detection; and
\item[\citet{Tsarapatsanis+2021}] discuss the importance of academic freedom, the diversity of legal and ethical norms, and the threat of moralism in the computational law field.
\end{description}

\section{Conclusion}
\label{conclusion}

This workshop is the fourth event from a series of workshops on automatic extraction of socio-political events from news, organized by the Emerging Market Welfare Project, with the support of the Joint Research Centre of the European Commission, with contributions from many other prominent scholars in this field. The purpose of this series of workshops is to foster research and development in the area of event extraction of socio-political events. 

The topics cover a wide range of applications and technologies: event detection via text classification, detection of news bias, fake news detection, modality analysis through syntactic parsing, event argument extraction and aggregation, a new geo-coding algorithm and the creation of a new geocoding dataset. Most of the papers are dedicated to protest events, one paper is about industrial reports, and one paper discusses generic events, not related to the socio-political topic.

The papers in this issue of the workshop make use of state-of-the-art NLP technologies, such as Deep Learning, Word Embeddings and Transformers. Most of the papers use the BERT model: some use the pre-trained existing models, others train domain-specific ones, and one of the paper introduces a modified version of BERT. Most papers use BERT embeddings as features in their models and one paper discusses an algorithm, which uses a full syntactic parser. Sentiment analysis is used in one paper, which studies the political bias of the news.

The shared task results shed light on critical aspects of the automated socio-political extraction and evaluation methodology.~\footnote{Detailed results are summarized in the overview papers~\citep{Hurriyetoglu+2021c,Haneczok+2021}, and evaluation (Task 3)~\citep{Giorgi+2021} of event databases. Moreover, system description papers provide full details} 

\section*{Acknowledgments}
The authors from Koc University were funded by the European Research Council (ERC) Starting Grant 714868 awarded to Dr. Erdem Y\"{o}r\"{u}k for his project Emerging Welfare.

\bibliographystyle{acl_natbib}
\bibliography{acl2021} % ,anthology,acl2021

%\appendix

\end{document}